\documentclass[a4paper,fleqn]{cas-sc}

\usepackage[numbers,sort&compress]{natbib}

\usepackage{graphicx}
\usepackage{subcaption}

\usepackage{tikz}
\usepackage{pgfplots}

\usepackage{multirow}
\usepackage{amssymb}
\usepackage{amsmath}

\usepackage{booktabs,arydshln}
\usepackage[export]{adjustbox}
\usepackage{float}

\usepackage[acronym]{glossaries}
\newacronym{cer}{CER}{Character Error Rate}
\newacronym{ocr}{OCR}{Optical Character Recognition}
\newacronym{htr}{HTR}{Handwritten Text Recognition}
\newacronym{cv}{CV}{Cross Validation}
\newacronym{dl}{DL}{Deep Learning}
\newacronym{petl}{PETL}{Parameter-Efficient Transfer Learning}
\newacronym{peft}{PEFT}{Parameter-Efficient Fine-Tuning}

\newacronym{icl}{ICL}{In-Context Learning}
\newacronym{rag}{RAG}{Retrieval-Augmented Generation}

\newacronym{id}{ID}{In-Domain}
\newacronym{cd}{CD}{Cross-Domain} 

\newglossaryentry{vlm}
{
  name={VLM},
  description={Vision-Language Model},
  first={Vision-Language Model (\glsentrytext{vlm})},
  plural={VLMs},
  descriptionplural={Vision-Language Models},
  firstplural={Vision-Language Models (\glsentryplural{vlm})}
}
\newglossaryentry{llm}
{
  name={LLM},
  description={Large Language Model},
  first={Large Language Model (\glsentrytext{llm})},
  plural={LLMs},
  descriptionplural={Large Language Models},
  firstplural={Large Language Models (\glsentryplural{llm})}
}
\newglossaryentry{crnn}
{
  name={CRNN},
  description={Convolutional Recurrent Neural Network},
  first={Convolutional Recurrent Neural Network (\glsentrytext{crnn})},
  plural={CRNNs},
  descriptionplural={Convolutional Recurrent Neural Networks},
  firstplural={Convolutional Recurrent Neural Networks (\glsentryplural{crnn})}
}

\newglossaryentry{lvlm}
{
  name={LVLM},
  description={Large Vision-Language Model},
  first={Large Vision-Language Model (\glsentrytext{lvlm})},
  plural={LVLMs},
  descriptionplural={Large Vision-Language Models},
  firstplural={Large Vision-Language Models (\glsentryplural{lvlm})}
}
\newacronym{lstm}{LSTM}{Long Short-Term Memory}

\makeatletter
\providecommand*{\input@path}{}
\g@addto@macro\input@path{{tables/}}
\makeatother
\graphicspath{{figures/}{images/}}

\begin{document}
\let\WriteBookmarks\relax
\def\floatpagepagefraction{1}
\def\textpagefraction{.001}

\shorttitle{Exploring In-Context Learning for Handwritten Text Recognition}

\shortauthors{Eric Ayllon et~al.}

\title [mode = title]{Exploring In-Context Learning for Handwritten Text Recognition}



%

\author[1]{Eric Ayllon}[orcid=0009-0005-7468-1712]

\cormark[1]


\ead{eric.ayllon@ua.es}


\credit{Methodology, Supervision, Formal analysis, Funding acquisition, Writing - Original Draft, Writing - Review \& Editing, Visualization}

\affiliation[1]{organization={University of Alicante},
            addressline={Carretera San Vicente del Raspeig s/n}, 
            city={San Vicente del Raspeig},
            postcode={03690}, 
            state={Spain},
            country={Spain}}

\author[1]{Abel Gandia}[orcid=0009-0006-3840-3480]



\ead{abel.gandia@ua.es}


\credit{Methodology, Software, Formal analysis}


\cortext[1]{Corresponding author}



\author[1]{Jorge Calvo-Zaragoza}[orcid=0000-0003-3183-2232]


\ead{jorge.calvo@ua.es}


\credit{Methodology, Writing - Review \& Editing, Supervision}



\begin{abstract}
%
\relax
\gls{htr} systems have become an indispensable tool for the digitization of historical documents. Not only do they cut down time and cost, but they also allow democratizing access and processing of their contents by generating their transcripts. However, literature in \gls{htr} currently focuses mostly on specialized models that require large amounts of annotated samples to achieve satisfactory performance.

We explore the use of In-Context Learning with pre-trained \glspl{vlm} to create a transcription pipeline without updating the model’s parameters. We then evaluate this pipeline across multiple collections and models, and demonstrate that general-purpose \glspl{vlm} can be effectively taught how to transcribe handwritten text from images. To assess how our observations may translate to practical applications, we evaluate the performance in a \gls{cd} scenario, where context examples are drawn from a different collection than the query image.

Results in both the controlled \gls{id} scenario and the realistic \gls{cd} scenario follow the same patterns. First, as context size grows, the error range is expected to narrow towards the average performance. Thus, larger context sizes sacrifice the performance of the oracle-best sampling for lower expected error rates.

The results obtained show that, without any parameter updates, this methodology has strong potential to compete with traditional \gls{htr} in the presence of domain shift. Moreover, we show and argue that some context samplings work better than others and suggest more effort should be put into finding an ideal sampling method in future work.
\end{abstract}

\begin{graphicalabstract}
\includegraphics[width=\textwidth]{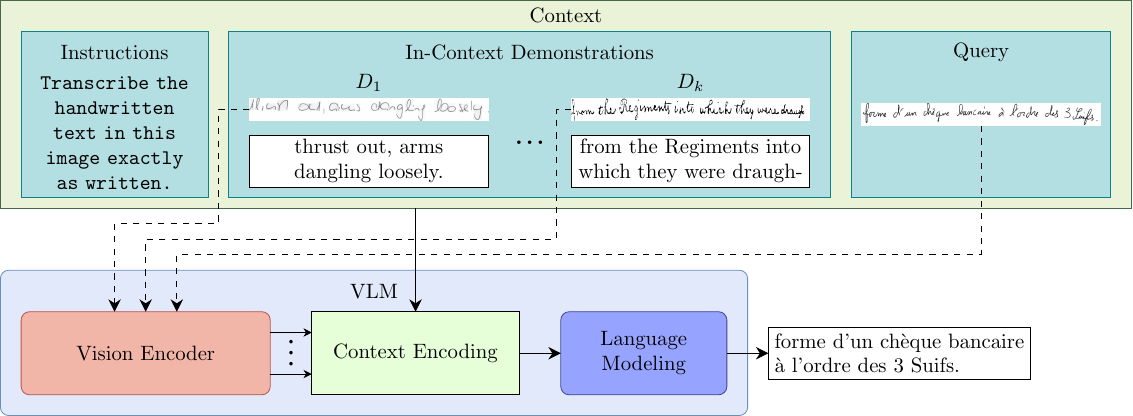}
\end{graphicalabstract}

\begin{highlights}
\item Training-free transfer learning via \acrshort{icl} for \acrshort{htr}.
\item Evaluation on multiple models and \acrshort{htr} collections.
\item \glsdesc{id} and \glsdesc{cd} evaluation.
\item Results with \acrshort{icl} are competitive with \acrshort{htr} literature.
\end{highlights}


\maketitle

\glsresetall
%
%
\section{Introduction}
The history and culture of our society have been recorded in manuscripts until the recent decades. While written documents may preserve their contents for a long time, physical storage media such as paper or even stone tablets degrade over time. To minimize information loss, original copies are often locked inside archives and libraries, and access to these is frequently restricted \cite{nikolaidou2022survey}. Modern technologies offer a more stable storage protocol based on storing digital copies in the form of images. However, merely capturing images of the documents only addresses one of the many problems regarding historical records: information lifespan. While digital images allow for long-term storage with little-to-no information loss, some documents benefit from other formats. Text documents, specifically, can be stored more efficiently if the textual contents of the image are stored instead of the original picture. Moreover, documents with highly degraded regions may be difficult to read while a clean transcription is not. Having access to the texts also allows machine-based indexing, searching, and processing of these documents, making them truly accessible to the public \cite{muehlberger2019transforming}.

While manual transcription of these historical records may seem a good idea, the sheer amount of documents waiting to be digitized or transcribed makes it unfeasible. Research and development of \gls{htr} pipelines aims to automate this process by creating systems capable of reading the contents of handwritten text images with minimal human intervention. Early formulations relied on heuristics and hand-crafted features, but the popularization of \gls{dl} and neural approaches allowed for more robust and adaptable systems \cite{htr-survey}. Over time, multi-step pipelines have been simplified, trading complexity and inter-module interactions for data-driven holistic solutions. Current literature in \gls{htr} is now divided into two approaches: segmentation-based and end-to-end. While both methodologies bring some benefits, the latter has become increasingly popular during the last decade \cite{htr-survey,DAN} thanks to the emergence of Vision-Language Modeling techniques, which enable enriching language modeling with visual information. Early proposals such as the one by \citet{DAN} focused only on fusing text and vision modalities via cross-attention in a transformer decoder, but recent works explore alternative formulations \cite{VLM-zeroshot-HTR,VLM-vietnamese-HTR} such as using \glspl{lvlm} \cite{VLM-Survey-2024}.

The data-driven nature of modern \gls{dl} approaches bounds inference performance to the similarity between training data and the target domain \cite{domain-adaptation}. \gls{htr} literature is no exception; inference quality under the effect of domain shift is subject to consistent detriment. As architectures become more complex and rely on fewer predetermined restrictions, transcription quality detriment due to insufficient or dissimilar training data becomes steeper \cite{HTR-ood}. However, the study by \citet{HTR-ood} on generalization of \gls{htr} architectures clarifies that not even the simplest models can achieve satisfactory transcription quality when their training data diverges too much from that of the target domain. Thus, an adaptation protocol prior to, or during, inference is critical to mitigate domain shift and unlock the full potential of \gls{dl} models.

The cost of training specialized systems from scratch, and then adapting them to a target domain, together with the recent trend of increasing model size, motivated the development of new \textit{open research} standards. To democratize access and use of these systems, the latest standards advocate for the publication of algorithms, data, and weights. This measure saves researchers and users a lot of time and resources while still allowing them to benefit from new technological advancements. However, these models, now called \textit{pre-trained} models, still need to be adapted to a target task or domain. While transfer learning demands fewer resources, the scale of many pre-trained models makes it prohibitively expensive. To palliate the effect of model size scaling on transfer learning costs, efforts on efficient transfer learning research have grown a lot recently, with special focus on parameter-efficient methods \cite{PETL}. Although the most popular class of transfer learning methods is those based on fine-tuning \cite{PEFT,efficient-VLMs}, the popularity of \glspl{llm} motivated the development of alternative means of adaptation other than updating the model's parameters. These alternative methods include: prompt-tuning \cite{prompt-tuning,prefix-tuning,p-tuning-v2}, optimizing the prompt; \gls{rag} \cite{RAG-best-practices,RAG-evaluation-survey,multipath-RAG}, enhancing the prompt with information from relevant documents; and \gls{icl} \cite{in-context-learning-survey,instruction-tuning-survey,ICL-what-matters-in-demonstrations,many-shot-ICL}, injecting examples into the prompt rather than training the model with them.


Despite the growing popularity of pre-trained \glspl{vlm} and the recent consideration of these architectures in \gls{htr} \cite{DAN,VLM-zeroshot-HTR}, their fitness for such a task still remains underexplored. Some works have already focused on specialized designs and training, while others evaluate pre-trained \glspl{vlm}. We find an important gap in the current literature, as there are no studies on efficient adaptation of \glspl{lvlm} to improve their performance on \gls{htr}. Based on the intrinsic ability of \glspl{vlm} to integrate vision and text inputs to perform complex vision tasks, and their emergent ability to learn from the context they are provided, we theorize that a transfer learning protocol based on \gls{icl} would be ideal for inference-time adaptation of these models. Given a handwritten text image for inference,
a set of example demonstrations of how the task should be resolved is given as context, thus helping the model understand the expected output structure and improving the transcription quality. The main objective of this study is, therefore, to verify whether \gls{icl} improves transcription quality. To properly assess how reliable this methodology can be and its true potential, we compare the transcription error with the best and worst context sets to the expected error. Moreover, we study the impact of \gls{icl} on transcription quality as the number of in-context examples, hereinafter called \textit{context size}, changes.

To test our hypothesis, we evaluate multiple pre-trained \glspl{vlm} on several \gls{htr} corpora of varying language and nature. For a thorough examination of the potential of \gls{icl} on this setup, we compare the transcriptions across multiple context samplings and sizes for each image in the evaluation test.
To complete the experimental pipeline, we divide our evaluation into two scenarios: \gls{id}, where context samples belong to the train split of the target corpora; and Cross-Dataset (or \gls{cd}), where context samples belong to the train set of any corpora except for the target domain.

Our contributions can be summarized as follows. First, we propose a hypothesis about the benefits of using \gls{icl} to improve the transcription quality obtained by prompting \glspl{vlm} to extract the handwritten text from images. Then, we study the conditions under which this hypothesis holds, reporting the results of our experiments and providing a detailed analysis of the observed trends and a discussion about their implications.

The remainder of this document is structured as follows: Section \ref{sec:background} contextualizes this work in the literature of \gls{htr} and introduces related works on foundation models, \glspl{vlm} and \gls{icl}. Section \ref{sec:methodology} describes the problem and defines relevant formulations for the experiments. Section \ref{sec:experimental-setup} details the setup and hyperparameters to facilitate replicating the experiments. Section \ref{sec:results} presents and analyzes the results of our experiments, providing relevant insights on their implications. Finally, Section \ref{sec:conclusions} summarizes the experiments, findings, and insights from this work, and proposes future research avenues on this matter.

\section{Background}\label{sec:background}
This work belongs in the intersection of three fields: \acrlong{htr}, Vision-Language Modeling, and \acrlong{icl}. In this section, we provide some context about them to help understand where this study lies within each field. Thus, we first provide a brief overview of how automated reading systems have been designed over the recent decades. Then, we provide some insights about the inspiration of models capable of fusing visual and textual information, which are called \glspl{vlm}. Finally, we present the concept of \gls{icl} and its evolution from an emergent property to a property that should be favored during training.

\subsection{Handwritten Text Recognition}
Historically, advancements in \gls{htr} have always been inspired by the development of similar fields, such as Automatic Speech Recognition (ASR) \cite{ASR-1,ASR-survey}, sequence modeling \cite{sequence-modeling}, and language modeling \cite{rnn-language-modeling,natural-language-generation-survey,LLM-survey}. Bringing many different approaches for this task, and various architectures following each approach \cite{htr-survey,HTR-onn}.

While state of the art in modern \gls{htr} architectures outperformed Hidden Markov Models a long time ago \cite{htr-survey}, current architectures still rely on a similar formulation. First, visual features are extracted from the input image and given to a sequence modeling module. When images contain singular lines of text, a bidirectional \gls{lstm}-based network is enough to model the output sequence from the feature map \cite{crnn-ctc-enrique_vidal}. For more complex cases such as paragraphs or even complete pages, a transformer decoder allows picking which visual features are relevant for the sequence modeling at each iteration \cite{DAN}. While the recurrent network approach directly works on the feature maps, the transformer approach combines sequence features with visual features through the cross-attention module.

\subsection{Vision-Language Models}
Motivated by the numerous milestones achieved by \glspl{llm} in language processing and modeling, the computer vision community developed an architecture that merges vision knowledge and language knowledge: the \glsdesc{vlm}. The idea behind this new paradigm-changing architecture is to enable general knowledge learned from text to be used in vision-heavy tasks \cite{VLM-Survey-2024,efficient-VLMs}.

Since the core of \glspl{vlm} lies in the integration of the vision and text modalities, extensive research has been made on modality fusion \cite{VLM-modality-fusion}. Even though no singular method stands out, the \gls{lvlm} approach seems to be gaining popularity due to its simplicity and compatibility with the already popular \gls{llm} paradigm \cite{LLM-review,VLM-Survey-2024,efficient-VLMs}.

While this is not the first attempt at using \glspl{vlm} for \gls{htr}, \citet{VLM-zeroshot-HTR} only covered the zero-shot performance of this architecture on the task. While their out-of-the-box performance can be competitive in some domains and tasks thanks to their general knowledge, the transcription quality is bound to be suboptimal without any kind of adaptation procedure. This work aims at bridging this literature gap.

\subsection{In-Context Learning}
As language models and their training data pools scale in size, they have started developing a powerful emergent ability. By definition, language models condition their output on a given context; the more this context is consulted for the output generation, the better \cite{instruction-tuning,knowledge-augmented-LLM}. Many of the latest models are even capable of following instructions provided in their context \cite{instruction-tuning-survey,visual-instruction-tuning}. By taking this instruction protocol one step further, some models are able to learn or improve their ability to perform downstream tasks without explicit specific training, just by taking some demonstrations of inputs and their expected outputs \cite{in-context-learning-survey}.

Recent studies on this paradigm have shown that, although it is interesting to work with such an emergent property of language modeling, this can also be a trainable skill. \citet{in-context-learning-survey} explain different strategies developed in the literature of \gls{icl}, these include alterations of the pre-training protocol \cite{pretraining-for-ICL} itself and fine-tuning for \gls{icl}. The latter is more widely adopted \cite{finetuning-no-tailored-data,finetuning-for-ICL-2,finetuning-for-ICL-3} mainly due to the lack of tailored data \cite{finetuning-no-tailored-data}, and it also allows improving \gls{icl} performance of already existing models as opposed to pre-training the whole models from scratch.

\section{Methodology}\label{sec:methodology}
This section is divided into three parts, each describing a different part of our methodology. First, we provide in Section \ref{sec:language-modeling} a brief and concise formulation for Language Modeling, the engine of our transcription pipeline. Then, in Section \ref{sec:htr-vlm} we explain the role of the vision encoder, and its interaction with the context encoding to inform the language model about written text information, allowing us to approach \gls{htr} with Vision-Language Modeling. Finally, we summarize in Section \ref{sec:icl} the typical \gls{icl} formulation and the role of context within our pipeline to enhance transcription quality without updating any model parameter.

\begin{figure*}[pos=ht] 
    \centering
    \includegraphics[width=1\linewidth]{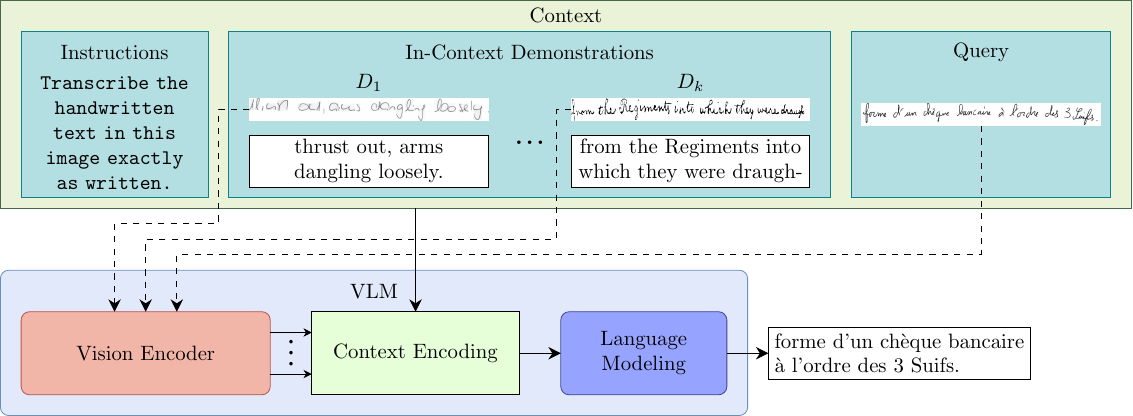}
    \caption{\acrlong{icl}-based adaptation protocol studied. The demonstrations from the context are randomly sampled for each query image.}
    \label{fig:methodology}
\end{figure*}

\subsection{Language Modeling}\label{sec:language-modeling} 
A language is a set of sequences composed of atomic elements, often called \textit{tokens}, from a vocabulary $\Sigma$ that satisfy some predicate $P$. Modern language modeling approximates $P$ with a probability function. A sequence, also called \textit{sentence}, $\mathbf{s} \in \Sigma^*$ is valid and belongs to a language $L$ if $P(\mathbf{s}) = 1$. To make this possible, the probabilistic approximation of $P$ rates the probability that a token $s_n$ follows a given context $s_1,...,s_{n-1}$:

\begin{equation}\label{eq:simplified-conditional-probability}
    P(\mathbf{s}) \approx p(s_i|s_1...s_{i-1})
\end{equation}





Equation \ref{eq:simplified-conditional-probability} represents the core idea of language modeling, upon which all modern approaches are based. A prominent application of this formulation is text generation \cite{LLM-review}, which can be approached by iteratively selecting the most likely word given the current context $\hat{s}_i = \arg\max_{s_i\in \Sigma} p(s_i|s_{i-c}...s_{i-1})$, starting the generation loop with a special symbol commonly termed the \textit{start-of-sequence} token.

The standard protocol for representing characters and words from $\Sigma$, adopted in the literature over the last decade, is to use a dense vector called \textit{embedding} to represent each element in $\Sigma$ \cite{dense-embeddings-1,dense-embeddings-2,dense-embeddings-3,dense-embeddings-4}. The values of such vectors are optimized with supervised learning, and they represent the meanings of words and characters.

Figure \ref{fig:methodology} illustrates the language model as the last module of our pipeline, which takes the encoded context information and generates the most likely sentence based on it. Note that we have purposefully elided the text embedding for clarity, but a text encoding module is implicit in the models used in our pipeline.


\subsection{Handwritten Text Recognition and Vision-Language Modeling}\label{sec:htr-vlm}
Transcribing the contents of an image can be mathematically defined as applying a mapping from an image space $\mathcal{X}$ to a text or sequence space $\Sigma^*$. This mapping $f:\mathcal{X}\rightarrow\Sigma^*$ is traditionally divided into two steps: visual feature extraction $\phi_\text{features}$ and sequence decoding $\phi_\text{sequence}$. While multi-step pipelines are usually more cumbersome due to the need to obtain each component separately, modern approaches allow training $\phi_\text{features}$ and $\phi_\text{sequence}$ simultaneously in an end-to-end training protocol. Formally, an \gls{htr} system aims at approximating $f$ as

\begin{equation}\label{eq:htr-original}
    f(x) \approx \phi(x) = \phi_\text{sequence} \circ \phi_\text{features} (x),
\end{equation}

\noindent
where the parameters of $\phi_\text{sequence} \circ \phi_\text{features}$ are optimized via supervised training.

For our pipeline, we reformulate the decoding step as a sequence modeling task. Thus, after obtaining dense vectors representing visual features extracted from the input image $\phi_\text{features} (x)$, a language model is asked to generate the most likely sentence conditioned by the visual features. The conditioning can be performed via cross-attention as proposed by \citet{DAN}, also considered by \citet{smt} in a similar field called Optical Music Recognition, using a context for text and another one for vision tokens; or directly embedding the visual features among the text embeddings in a singular multi-modal context. The choice of modality fusion, a more popular term for the conditioning, depends on the model used, although neither has been proved to consistently outperform the other \cite{VLM-modality-fusion}.

The diagram in Figure \ref{fig:methodology} represents $\phi_\text{features}$ as the "Vision Encoder" module, and the context encoding receives information both from text (directly from the context) and vision (output of the vision encoder). While the diagram is agnostic to the modality fusion type, most of the models we use follow the \gls{lvlm} (shared context) approach.

\subsection{In-Context Learning}\label{sec:icl}
The idea behind \gls{icl} is quite similar to how supervised training works. However, \gls{icl} aims at teaching the model to perform a specific task without updating any of its parameters. This is possible thanks to an emergent property of \glspl{llm}. Although this property is directly inherited from the formulation of language modeling presented in Section \ref{sec:language-modeling}, only large-scale models have shown the ability to exploit the context this way.

To achieve this goal, a language model is provided with an instruction to clarify the objective task. This already conditions the text that is going to be generated, as the history of \textit{events} contains the instruction; then a set of input-output pairs, often called \textit{demonstrations} \cite{in-context-learning-survey}, are provided to further enrich the context. Finally, the query input is appended to the context, and the model is then expected to replicate the structure of its context by generating the most likely output for the query.

In our pipeline, the input-output pairs are images and their transcriptions (See "In-Context Demonstrations" block in Figure \ref{fig:methodology}), and the query is an image (See "Query" in Figure \ref{fig:methodology}). To construct the context embeddings, a vision encoder extracts appropriate representations of the context and query images, as we have explained in \ref{sec:htr-vlm}, and combines these representations with the text embeddings either by concatenation or by feeding them to the cross-attention of the language model.

For closure, we formulate this process mathematically by combining the equations we have presented so far. Given the query image ($x_q$) and the context $C=\{(x_i,l(x_i))\}_{1}^{k}$ containing $k$ images and their ground truth transcription $l(x)$, we reformulate equation \ref{eq:htr-original} as:

\begin{equation}\label{eq:htr-icl}
    f(x_q) \approx \phi(x_q, C) = \phi_\text{sequence} \circ \phi_\text{features} (x_q, C)
\end{equation}

\section{Experimental setup}\label{sec:experimental-setup}

\subsection{Corpora}\label{sec:corpora}
To assess the performance of the presented methodology, we considered using six corpora of varying characteristics in our experiments. Specifically, we used two English datasets: IAM \cite{IAM}, a multi-writer database; and Washington \cite{Washington}, a collection of handwritten correspondence letters from the 18th century by George Washington. Rimes-2011 \cite{Rimes-2011}, a French handwritten correspondence multi-writer database, where correspondence letters were written by volunteers to fictitious companies. ICFHR2016 \cite{ICFHR2016}, a dataset based on the competition on \gls{htr} held in the International Conference on Frontiers in Handwriting Recognition (ICFHR) in 2016. The ICFHR2016 collection gathers historical handwritten documents from the Ratsprotokolle collection composed of minutes of the council meetings held from 1470 to 1805 in early modern German. The Ludovico Antonio Muratori (LAM) dataset \cite{LAM} contains scans of handwritten documents and correspondence authored by the Italian historian L. A. Muratori over the course of 60 years. Finally, the SaintGall dataset \cite{SaintGall} contains the hagiography \textit{Vita sancti Galli} by Walafrid Strabo written in the 9th century in Latin. Table \ref{tab:corpora} summarizes the most important details about these collections, including the language and the total number of samples. For reproducibility's sake, the data splits used correspond to the original data splits, including the cross-validation for the Washington dataset.

{
\setlength{\intextsep}{0.25cm}
\begin{table*}[ht]
    \centering
    \caption{Language and total number of images in each collection.}
    \label{tab:corpora}
    \begin{tabular}{l*{6}{c}}
        \toprule
        Corpus && Language && Total samples && \\
        \midrule
        LAM &\phantom{--}& Italian &\phantom{--}& 23353 && \includegraphics[width=.49\textwidth,valign=m]{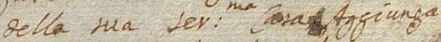} \\
        SaintGall && Latin && 1175 && \includegraphics[width=.49\textwidth,valign=m]{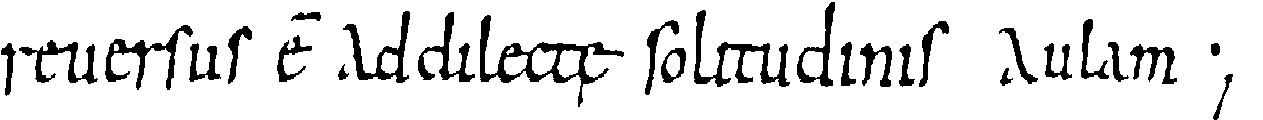} \\
        Washington && English && 656 && \includegraphics[width=.49\textwidth,valign=m]{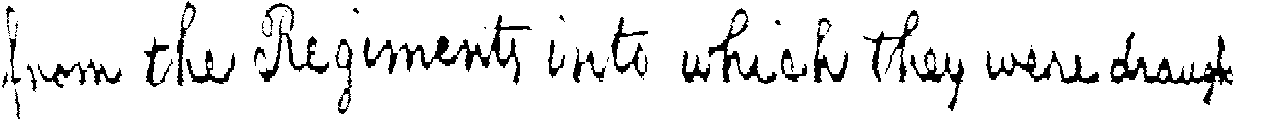} \\
        IAM && English && 8022 && \includegraphics[width=.49\textwidth,valign=m]{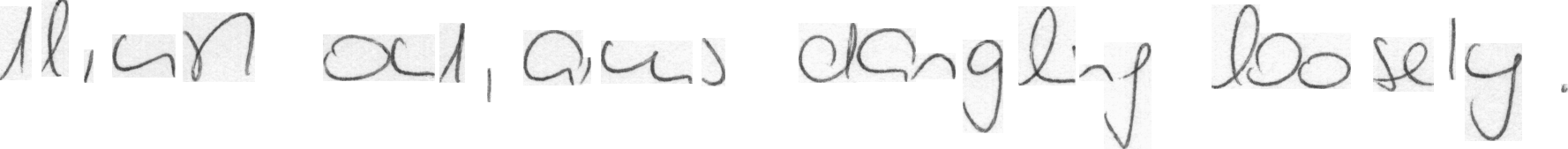} \\
        Rimes-2011 && French && 10966 && \includegraphics[width=.49\textwidth,valign=m]{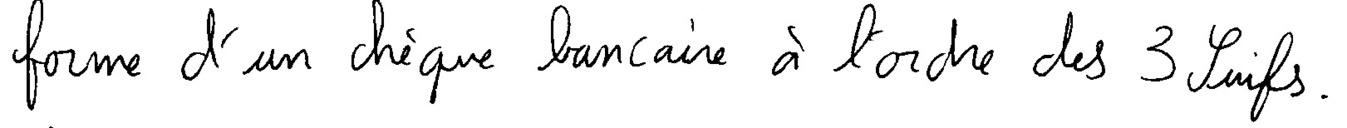} \\
        ICFHR2016 && German && 9504 && \includegraphics[width=.49\textwidth,valign=m]{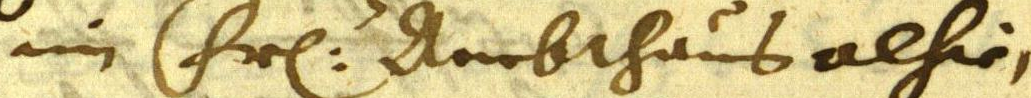} \\
        \bottomrule
    \end{tabular}
\end{table*}
}

\subsection{Metrics}\label{sec:metrics}
For the evaluation of the transcription quality, we use the \gls{cer}, a standard evaluation metric that is also widely used in the literature of \gls{htr} \cite{htr-survey,htr-omr-calibration}. This metric is defined as the Levenshtein edit distance \cite{edit-distance}, divided by the length of the ground truth label. It serves as an estimation of the effort of an expert to correct a predicted transcription. If we represent the ground truth transcription of a given sample $x$ as $l(x)$, the \gls{cer} is defined as follows:

\begin{equation}\label{eq:cer}
    \text{CER}(\phi(x, C), l(x)) = \frac{\text{edit distance}(\phi(x, C), l(x))}{|l(x)|}
\end{equation}

Since the predicted transcription depends on the context ($\phi(x, C)$), we also study the error range, i.e., the difference between the minimum and maximum \gls{cer} for the same sample using different contexts of the same size.

\subsection{Experimental scenarios}\label{sec:scenarios}
In our experimental setup, we evaluate the pipeline on each collection by sampling contexts from its training split, following the typical supervised training methodology. By performing inference on a set of images similar to the context $C$, ideally independent and identically distributed to $C$, the knowledge gained from the demonstrations may translate better to the target image.

While this setting is useful for analyzing the potential of our method, it fails to evaluate it under similar conditions of real applications. Given a new collection with no annotations, the likelihood of having transcription demonstrations that are identically distributed to the new corpus is quite low. To properly assess the pipeline, we repeat the experiments with an additional restriction: context demonstrations must belong to a different corpus. To clarify the difference, while in the first scenario, context is drawn from the same collection, in this one context is drawn from any collection except for the corpus of the inference sample. In any case, context samples are only drawn from training splits, regardless of the corpus they belong to. From now on, the first scenario will be referred to as \gls{id}, and the second one will be called \gls{cd}.

\subsection{Implementation details}\label{sec:implementation}
To make sure the experiments are completely replicable, we have gathered all hyperparameters and details about the experiments that remain untold in this section. The models we have evaluated can be found in Table \ref{tab:models}, where their HuggingFace ID is reported along with their short name.

{
\setlength{\intextsep}{0.25cm}
\begin{table}[ht]
    \centering
    \caption{Models used in the experiments, along with their Huggingface ID
    .
    }
    \label{tab:models}
    \begin{tabular}{l*{2}{l}}
        \toprule
        Model & Huggingface ID \\
        \midrule
        Qwen2.5-VL \cite{Qwen2VL} & \texttt{Qwen/Qwen2.5-VL-7B-Instruct} \\ 
        Qwen3-VL \cite{Qwen3} & \texttt{Qwen/Qwen3-VL-8B-Instruct} \\ 
        Gemma 4 & \texttt{google/gemma-4-E4B-it} \\ 
        Kimi \cite{Kimi} & \texttt{moonshotai/Kimi-VL-A3B-Instruct} \\ 
        \bottomrule
    \end{tabular}
\end{table}
}

The context sizes we considered in our experiments range from 0 (baseline, no \gls{icl}) to 16 using powers of 2: 1, 2, 4, 8, 16. We stopped at 16 based on preliminary results, which showed marginal improvements in transcription quality; thus, contexts larger than 16 are not expected to justify the increase in computation. Each evaluation is repeated with multiple different context samplings to estimate the expected error due to the effect of random sampling, as well as empirical proxies for the upper and lower bounds. Specifically, we report min. \gls{cer} as the estimation of the performance achievable under favorable context selection, while max. \gls{cer} is the opposite, i.e., attainable performance under unfavorable context selection. The number of repetitions performed is always fixed to 30 due to computation constraints. For context size of 0, though, just one repetition is enough since the result should not be significantly different across multiple executions.

The experiments were implemented using Python's \texttt{vllm} module. This library allows using a wide range of models available at HuggingFace through the transformers library.

\section{Results}\label{sec:results}
To provide a thorough analysis of the results of our experiments, we organize this section as follows. Before the analysis of the results, we explain how the presented values are obtained. Then, for the analysis, we first present a summary of the results in the first scenario, where the potential domain shift is limited. By complementing the report with a detailed discussion about the observations we can make from the results, we aim to clarify the implications of this methodology. Then, we compare the results from \gls{id} to those of the \gls{cd} scenario. This allows studying how robust the adaptation protocol is towards domain shift dimensions such as change in paper type and background, different language and calligraphy, and age or date of writing.

The overview of our results follows a simple structure. For each context size, we report the minimum, average, and maximum global \gls{cer}, thus the metric is calculated across the whole test set. These values are obtained as follows: for each inference image, we find the min., average, and max. edit distance across repetitions. Then, we sum the ground truth label length across samples from the test split, and we do the same for the minimum, mean, and maximum edit distances. Then, we divide the edit distances by the sum of the lengths to obtain the min., avg., and max. global \gls{cer} for a configuration (dataset, model, context size). To improve the clarity of the analysis and the results, we summarize them by averaging across models. For the complete table of results, we refer the reader to Appendix \ref{sec:complete-results}, where both \gls{id} and \gls{cd} results are gathered without averaging across models. Note that, despite different models achieving different performances, all of them follow the same trends we observe and discuss in this section.

{
\setlength{\intextsep}{0.25cm}
\begin{table}[hbt]
    \ifx\ifdoublecolumnsquish\undefined
        \setlength{\tabcolsep}{4pt} 
    \else
        \setlength{\tabcolsep}{1pt} 
    \fi
    \centering
    \caption{\gls{cer} (\%) per configuration in the \gls{id} scenario averaged across models.}
    \label{tab:res-id-summary}
    \begin{tabular}{*{2}{l}*{6}{c}}
        \toprule%
        \multirow{2}{*}{Collection} && \multicolumn{6}{c}{Context Size} \\%
        \cmidrule{3-8}%
         & & 0 & 1 & 2 & 4 & 8 & 16 \\%
        \midrule%
\multirow{3}{*}{IAM} & max & 17.68 & 24.20 & 17.28 & 14.95 & 12.72 & 9.61 \\
 & avg & 17.68 & 9.96 & 8.81 & 7.87 & 6.61 & 5.36 \\
 & min & 17.68 & 5.00 & 4.50 & 3.89 & 3.41 & 3.12 \\
\midrule
\multirow{3}{*}{ICFHR16} & max & 378.13 & 230.41 & 188.46 & 169.24 & 168.43 & 154.72 \\
 & avg & 378.13 & 88.24 & 82.79 & 80.61 & 78.87 & 77.47 \\
 & min & 378.13 & 61.16 & 60.26 & 59.35 & 57.92 & 56.93 \\
\midrule
\multirow{3}{*}{LAM} & max & 30.52 & 86.03 & 76.77 & 72.75 & 71.72 & 64.28 \\
 & avg & 30.52 & 25.16 & 23.46 & 22.27 & 21.63 & 21.02 \\
 & min & 30.52 & 12.73 & 12.18 & 11.73 & 11.21 & 10.84 \\
\midrule
\multirow{3}{*}{RIMES} & max & 8.81 & 31.69 & 28.15 & 18.44 & 16.93 & 12.35 \\
 & avg & 8.81 & 8.31 & 7.35 & 6.59 & 6.04 & 5.56 \\
 & min & 8.81 & 3.59 & 3.29 & 3.16 & 2.93 & 2.82 \\
\midrule
\multirow{3}{*}{SaintGall} & max & 32.98 & 28.17 & 28.86 & 31.21 & 31.65 & 27.74 \\
 & avg & 32.98 & 18.34 & 17.99 & 18.27 & 18.59 & 18.54 \\
 & min & 32.98 & 13.28 & 12.60 & 12.30 & 12.77 & 13.04 \\
\midrule
\multirow{3}{*}{Washington} & max & 14.98 & 18.53 & 17.98 & 15.61 & 14.91 & 13.14 \\
 & avg & 14.98 & 9.32 & 8.66 & 8.19 & 7.79 & 7.55 \\
 & min & 14.98 & 5.57 & 5.08 & 4.80 & 4.60 & 4.35 \\
        \bottomrule%
    \end{tabular}%
\end{table}
}

The results presented in Table \ref{tab:res-id-summary} prove that providing examples of how the target task should be solved affects the transcription quality. Moreover, while context size does generally allow for better results on average and for the worst-case proxy, its effect on the lower-bound proxy is greatly diminished. We observe that large contexts often reduce the maximum error considerably. Thus, if we consider the maximum \gls{cer} to be an empirical proxy for robustness, large contexts allow for more robust transcription pipelines.
The average \gls{cer} frequently benefits from larger contexts too, even though the performance gap is much smaller. Therefore, large contexts allow for higher expected transcription quality.
We believe it is important to note that most of the improvement obtained is already obtained in context sizes as small as 4, or even 2. Such is the case especially for SaintGall, which displays its peak performance at a context size of 2.
Finally, the minimum \gls{cer} shows minimal improvements when using large contexts, and in some cases even shows detriment instead of benefit. Considering this to represent the \textit{potential} of the pipeline, larger context sizes do not ensure optimal quality can be achieved.
Regardless of the context size, the results we present suggest \gls{icl} generally improves the transcription quality for all collections when the domain shift of the query input is limited.

{
\setlength{\intextsep}{0.5cm}
\begin{table}[htb]
    \ifx\ifdoublecolumnsquish\undefined
        \setlength{\tabcolsep}{4pt} 
    \else
        \setlength{\tabcolsep}{1pt} 
    \fi
    \centering
    \caption{\gls{cer} (\%) per configuration in the \gls{cd} scenario averaged across models.}
    \label{tab:res-ood-summary}
    \begin{tabular}{*{2}{l}*{6}{c}}
        \toprule
        \multirow{2}{*}{Collection} && \multicolumn{6}{c}{Context Size} \\
        \cmidrule{3-8}
         &  & 0 & 1 & 2 & 4 & 8 & 16 \\
        \midrule
\multirow{3}{*}{IAM} & max & 17.68 & 14.94 & 12.70 & 9.79 & 9.49 & 9.60 \\
 & avg & 17.68 & 7.73 & 7.25 & 6.91 & 6.85 & 6.82 \\
 & min & 17.68 & 5.55 & 5.56 & 5.62 & 5.63 & 5.62 \\
\midrule
\multirow{3}{*}{ICFHR16} & max & 378.13 & 832.93 & 686.98 & 602.99 & 618.41 & 441.28 \\
 & avg & 378.13 & 161.75 & 128.02 & 119.08 & 116.14 & 103.73 \\
 & min & 378.13 & 68.37 & 65.55 & 65.22 & 64.86 & 65.19 \\
\midrule
\multirow{3}{*}{LAM} & max & 30.52 & 94.39 & 90.62 & 84.94 & 78.12 & 75.96 \\
 & avg & 30.52 & 26.48 & 25.71 & 24.95 & 24.20 & 24.18 \\
 & min & 30.52 & 13.25 & 13.08 & 13.30 & 13.29 & 13.33 \\
\midrule
\multirow{3}{*}{RIMES} & max & 8.81 & 33.97 & 25.08 & 21.46 & 15.68 & 14.57 \\
 & avg & 8.81 & 8.92 & 8.22 & 7.67 & 7.31 & 7.12 \\
 & min & 8.81 & 3.95 & 3.89 & 4.01 & 3.98 & 4.01 \\
\midrule
\multirow{3}{*}{SaintGall} & max & 32.98 & 57.68 & 48.45 & 40.85 & 44.08 & 37.86 \\
 & avg & 32.98 & 21.26 & 20.80 & 20.57 & 20.77 & 20.38 \\
 & min & 32.98 & 14.50 & 14.37 & 14.59 & 14.72 & 14.84 \\
\midrule
\multirow{3}{*}{Washington} & max & 14.98 & 37.90 & 24.38 & 17.74 & 20.54 & 17.58 \\
 & avg & 14.98 & 11.74 & 9.87 & 9.79 & 9.81 & 9.74 \\
 & min & 14.98 & 6.10 & 6.08 & 6.14 & 6.12 & 6.23 \\
        \bottomrule
    \end{tabular}
\end{table}
}

Let us now compare our previous observations to the \gls{cd} scenario. Table \ref{tab:res-ood-summary} presents the results from this scenario with the same structure and protocol for calculating the reported values as for Table \ref{tab:res-id-summary}.
Note that the baseline results (context size of 0) are the same as from the first scenario since having no context demonstrations results in zero-shot evaluations regardless.

In this scenario, we can see that transcription quality does improve with respect to the baseline. However, similar to SaintGall in the \gls{id} scenario, increasing the context size in this scenario does not always warrant an improvement with respect to the previous context size. We observe, however, a similar phenomenon regarding the extreme cases (best and worst). Thus, larger context sizes still allow for more stable transcription pipelines in this scenario, at the cost of potential, i.e., worse best-case observations.

While this strategy provides moderately low-error transcriptions in most collections, the range of error rates achieved for ICFHR16 is considerably higher than the ranges of other collections. Having this in mind, we highlight this collection as a serious challenge for this methodology. This suggests a pipeline obtained by this procedure may not perform well regardless of the characteristics of the images to transcribe and their contents. This is not exclusive to the \gls{cd} scenario, we observe the same phenomenon in the \gls{id} scenario too.

\subsection{Qualitative analysis}
While transcription quality in general is reasonable, the worst-case scenario reports very high error rates. After a detailed analysis of the predictions made by different \glspl{vlm}, we have identified multiple types of failed transcriptions.

When the query image is allegedly ``too difficult'' for the model to transcribe, a lengthy explanation of why it is hard to transcribe is produced instead. This has been observed on short text images with individual letters, slightly blurry images, or even sequences of perfectly readable letters that do not compose real words. Moreover, this phenomenon is not consistent, as some contexts for the same query allow the models to transcribe the contents of the image, or at least make a guess without a verbose answer. We classify these erroneous responses as \textit{task failure} since the model does not even provide an answer following the context structure, even if it does recognize the text in the image.

Another type of error we observe is a hallucination during language generation. The generation loop starts repeating a part of the predicted text indefinitely. While the task structure is followed, the language model fails to ensure a comprehensive string is generated; thus, we classify this error as \textit{language failure}.

The last error type we have found is the \textit{transcription failure}, which would ideally be the only type of error we find. This arises when the structure of the context is replicated in the generated text, i.e., no needless verbosity, and the prediction does not contain an infinite loop of meaningless text.

\subsection{Impact of model choice}
\begin{figure*}[pos=ht]
    \centering
    \includegraphics[width=0.99\linewidth]{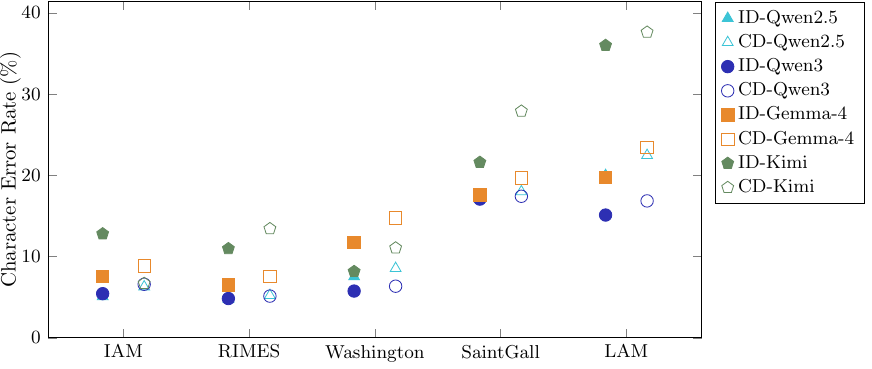}
    \caption{Comparison of transcription quality across different models on each collection. The plot should be read collection by collection. For each collection, we present the average \gls{cer} averaged across context sizes; for \gls{id} (filled marks) and \gls{cd} (outline marks). In ICFHR16, the average \gls{cer} obtained for each configuration in the same order are: for \gls{id} 73.15,82.95,85.35, and 84.93; and for \gls{cd} 75.41,96.38,149.07, and 182.12.}
    \label{fig:vlm-comparison}
\end{figure*}

Despite the promising results we have presented so far, the final pipeline's performance depends heavily on the underlying \gls{vlm} used for the transcription. Thus, we provide a comparison of the average \gls{cer} obtained by each model for each collection. Figure \ref{fig:vlm-comparison} displays the average \gls{cer} across context sizes greater than 0 for both scenarios, \gls{id} and \gls{cd}. Since the error range in ICFHR16 is completely different from other collections, it has been excluded from the plot for clarity. However, the \gls{cer} obtained in this dataset is provided in the caption.

A glance at Figure \ref{fig:vlm-comparison} is enough to conclude Qwen3 takes the lead with remarkable consistency in both scenarios. The only exceptions are IAM and ICFHR16. However, the performance gap in IAM is negligible, and both models struggle a lot in ICFHR16. While this figure proves the performance is bound by the model used for transcription, the success of the pipeline does not rely on using a specific underlying \gls{vlm}.

\subsection{Comparison to current literature}
To contextualize our results, we have collected the \gls{cer} reported in recent works studying the performance and generalization of \gls{htr}-specific architectures. Specifically, we focus on the best performance reported for each dataset used in our experiments. Note that the models reported as best performing for each collection have been specifically designed for, and trained on \gls{htr} tasks. This methodology, however, requires no parameter updates at all in either of the scenarios we have studied, but they are pre-trained models that already come with a massive pre-training. We have excluded the LAM dataset from this comparison because we could not find updated \gls{id} results or even \gls{cd} performance evaluations in the literature of \gls{htr}. Moreover, to simplify the analysis, we focus on comparing the results we obtain using Qwen3 only, as it is the best-performing \gls{vlm} in most of our experiments.

\begin{figure*}[pos=h]
    \centering
    \includegraphics[width=0.99\linewidth]{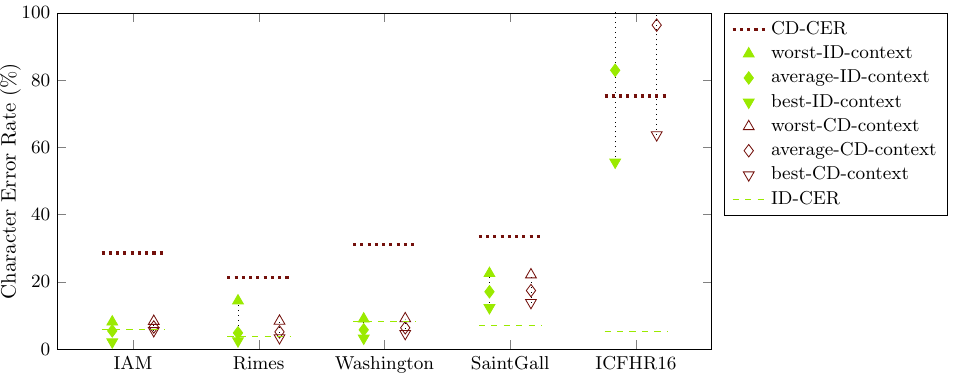}
    \caption{Comparison of our proposal to current literature in the collections used in our experiments. The reported values for our pipeline correspond to Qwen3. For each collection, we present the \gls{cer} obtained with the absolute worst \protect\tikz{\protect\path plot[only marks, mark=triangle,mark options={mark size=3.5pt}] coordinates {(0,0)};}, average \protect\tikz{\protect\path plot[only marks, mark=diamond,mark options={mark size=3.5pt}] coordinates {(0,0)};}, and absolute best \protect\tikz{\protect\path plot[only marks, mark=triangle, mark options={mark size=3.5pt,rotate=180}] coordinates {(0,0)};} contexts; for \gls{id} (filled in green) and \gls{cd} (only outline in brown). The vertical dotted lines group results of the same scenario, and the horizontal dashed and dotted lines represent the performance reported from \gls{htr} literature on each dataset. The literature results were collected from \cite{HTR-ood}.}
    \label{fig:sota-comparison}
\end{figure*}

Figure \ref{fig:sota-comparison} provides an overview of our results in an even more summarized way than Tables \ref{tab:res-id-summary} and \ref{tab:res-ood-summary}. Specifically, for the minimum and maximum \gls{cer}, the lowest (and highest, respectively) reported value from the summary tables is selected to represent the absolute best and absolute worst contexts regardless of context size, excluding size 0. For the average performance, the plotted value corresponds to the average across context sizes greater than 0. \gls{id} and \gls{cd} performance have been presented independently (filled marks vs outline marks), and vertical dotted lines have been added to group results from the same scenario. To contextualize within the literature of \gls{htr} the performances we obtained, we have plotted the best performance, both \gls{id} and \gls{cd}, obtained in each collection in the work by \citet{HTR-ood}.

There are multiple insights we can extract from Figure \ref{fig:sota-comparison}. First, in most collections, our average \gls{id} performance is quite similar to theirs, thus improving upon their \gls{cd} performance. Moreover, while performance in \gls{id} is always better than in the \gls{cd} scenario, our \gls{cd} transcription quality is competitive with \gls{htr}-specific systems in these settings. The implications of this are clear: a practical application using In-Context Learning would have a notable advantage over current \gls{htr}-specific models when training or fine-tuning the system is not feasible.
Another important detail to note is the case of ICFHR16, which, as previously noted, poses a serious challenge for this adaptation strategy. We believe this may be due to a linguistic divergence between early modern German and contemporary German. Thus, even if these general-purpose \glspl{lvlm} have been exposed to German script at any point of their pretraining, it may not completely align with the implicit language model in this collection.



\section{Conclusions}\label{sec:conclusions}
\glsreset{vlm}
\glsreset{htr}
\glsreset{rag}
\glsreset{icl}
\glsreset{id}
\glsreset{cd}
This work presents an exploratory study on the use of pre-trained general-purpose \glspl{vlm} for \gls{htr}. We adapt the models to the task and domain with \gls{icl} and compare the performance to zero-shot evaluation. Our objective was to assess the sensitivity of transcription quality to context and context size. To do so, we evaluated the proposal under multiple sampling repetitions to study the effects of different samplings, and performed an empirical study of the minimum, average, and maximum \gls{cer} that can be achieved. Our formulation followed the typical \gls{icl} pipeline on 6 collections of 5 different languages. To study how this methodology may translate to a real application, we consider two scenarios: \gls{id}, where demonstrations belong to the same collection, and \gls{cd}, in which demonstrations are sampled from any collection but the one the query input belongs to.

After a thorough analysis of the results, we clearly see that the methodology we have studied is a promising parameter-free adaptation mechanism for \gls{htr}, achieving visible improvements over the baseline in both scenarios. Moreover, we find that context selection plays a crucial role in the adaptation. However, while larger context sizes allow for a more stable approach, most of the improvement is already obtained with contexts as small as 4 or even 2 demonstrations. Thus, we encourage future work to be made towards finding optimal demonstration selection methods.

After comparing our results with \gls{htr}-specific methodology, we conclude this framework is a good candidate for low-resource projects, where training or fine-tuning specialized systems is not possible. However, the strength of its benefits is not universal and strongly depends on the underlying \gls{vlm}, target collection, and context selection.

Apart from finding the ideal selection mechanism, future work on this topic should study the effect of language disparity across demonstrations, between prompt and context demonstrations, and between model and target collection. As well as studying the effect of demonstrations and context size in the distribution of failure types.

\printcredits

\bibliographystyle{model1-num-names}

\bibliography{cas-refs}



\appendix

\section{Complete results}\label{sec:complete-results}
This appendix presents the complete tables of results of our experimentation, for each model, collection, and context size. Table \ref{tab:res-id-complete} displays the results from the \gls{id} scenario, while Table \ref{tab:res-ood-complete} showcases the \gls{cd} performances.

Tables in this appendix display the \gls{cer} statistics for each collection and each context size, without averaging across models. These values are obtained with the following procedure: for each inference image, we find the min., average, and max. edit distance across repetitions. Then, we sum the ground truth label length across samples from the test split, and we do the same for the minimum, mean, and maximum edit distances. Then, we divide the edit distances by the sum of the lengths to obtain the min., avg., and max. \gls{cer} for a configuration (dataset, model, context size).

We refer the reader to the main text for a detailed analysis, since we observe the same trend for all models as in the summarized tables (the ones in the Results section of the main text). However, we provide a summary of the observations here. Despite all models following the same trend, we believe it is important to also take a look at each of them individually since models obtaining higher \gls{cer} values increase the average used in the summarized version. This, in turn, makes it seem this methodology is consistently worse than it is for some models and datasets. In fact, some of them even outperform specialized \gls{htr} models.

Both maximum and average \gls{cer} tend to improve as context size grows larger. Minimum \gls{cer} stays around similar values, but it also shows signs of performance drops for larger contexts in some configurations. Moreover, average \gls{cer} in the \gls{cd} scenario shows much less improvement when increasing context size.

{
\setlength{\intextsep}{0.25cm}
\setlength{\textfloatsep}{0.25cm}
\setlength{\floatsep}{0.25cm}
\begin{table*}[hb]
    \setlength{\tabcolsep}{4pt}
    \centering
    \caption{\gls{cer} (\%) per configuration in the \gls{id} scenario for each model.}
    \label{tab:res-id-complete}
    \begin{tabular}{*{3}{l}*{6}{c}}
        \toprule
        \multirow{2}{*}{Collection} &\multirow{2}{*}{Model}&& \multicolumn{6}{c}{Context Size} \\
        \cmidrule{4-9}
         &  &  & 0 & 1 & 2 & 4 & 8 & 16 \\
    \midrule
\multirow{12}{*}{IAM} & \multirow{3}{*}{Kimi-VL-A3B-Instruct} & max & 48.66 & 69.49 & 41.75 & 32.88 & 22.75 & 13.49 \\
 &  & avg & 48.66 & 19.40 & 15.41 & 12.64 & 9.56 & 7.06 \\
 &  & min & 48.66 & 5.96 & 5.68 & 5.14 & 4.94 & 4.59 \\
\cmidrule{2-9}
 & \multirow{3}{*}{Qwen2.5-VL-7B-Instruct} & max & 6.69 & 8.07 & 8.06 & 8.03 & 8.89 & 7.53 \\
 &  & avg & 6.69 & 6.05 & 5.84 & 5.47 & 4.52 & 3.51 \\
 &  & min & 6.69 & 4.02 & 3.42 & 2.65 & 1.96 & 1.71 \\
\cmidrule{2-9}
 & \multirow{3}{*}{Qwen3-VL-8B-Instruct} & max & 6.82 & 8.11 & 8.11 & 7.84 & 8.08 & 7.39 \\
 &  & avg & 6.82 & 6.24 & 6.06 & 5.71 & 5.07 & 4.14 \\
 &  & min & 6.82 & 4.38 & 3.72 & 3.08 & 2.41 & 2.14 \\
\cmidrule{2-9}
 & \multirow{3}{*}{gemma-4-E4B-it} & max & 8.56 & 11.13 & 11.18 & 11.04 & 11.16 & 10.04 \\
 &  & avg & 8.56 & 8.15 & 7.93 & 7.64 & 7.27 & 6.74 \\
 &  & min & 8.56 & 5.65 & 5.15 & 4.69 & 4.32 & 4.04 \\
\midrule
\multirow{12}{*}{ICFHR16} & \multirow{3}{*}{Kimi-VL-A3B-Instruct} & max & 896.86 & 360.24 & 266.68 & 208.34 & 187.78 & 171.94 \\
 &  & avg & 896.86 & 91.02 & 86.98 & 83.33 & 81.96 & 81.36 \\
 &  & min & 896.86 & 61.06 & 60.61 & 60.10 & 59.71 & 58.75 \\
\cmidrule{2-9}
 & \multirow{3}{*}{Qwen2.5-VL-7B-Instruct} & max & 76.89 & 156.54 & 126.29 & 139.52 & 146.38 & 143.68 \\
 &  & avg & 76.89 & 75.55 & 72.71 & 72.85 & 72.46 & 72.20 \\
 &  & min & 76.89 & 56.98 & 56.23 & 55.06 & 53.08 & 52.17 \\
\cmidrule{2-9}
 & \multirow{3}{*}{Qwen3-VL-8B-Instruct} & max & 80.95 & 182.14 & 210.43 & 199.95 & 223.59 & 194.05 \\
 &  & avg & 80.95 & 86.01 & 84.89 & 83.10 & 81.96 & 78.78 \\
 &  & min & 80.95 & 61.82 & 60.53 & 58.89 & 57.12 & 55.61 \\
\cmidrule{2-9}
 & \multirow{3}{*}{gemma-4-E4B-it} & max & 457.82 & 222.73 & 150.42 & 129.15 & 115.96 & 109.20 \\
 &  & avg & 457.82 & 100.36 & 86.58 & 83.16 & 79.10 & 77.54 \\
 &  & min & 457.82 & 64.78 & 63.68 & 63.35 & 61.77 & 61.19 \\
\midrule
\multirow{12}{*}{LAM} & \multirow{3}{*}{Kimi-VL-A3B-Instruct} & max & 50.26 & 206.15 & 185.80 & 182.19 & 175.07 & 154.16 \\
 &  & avg & 50.26 & 38.68 & 37.23 & 35.85 & 34.72 & 33.52 \\
 &  & min & 50.26 & 14.74 & 14.44 & 14.24 & 14.20 & 14.01 \\
\cmidrule{2-9}
 & \multirow{3}{*}{Qwen2.5-VL-7B-Instruct} & max & 23.34 & 72.81 & 57.60 & 47.10 & 57.34 & 52.03 \\
 &  & avg & 23.34 & 22.60 & 19.88 & 19.00 & 19.55 & 18.99 \\
 &  & min & 23.34 & 11.97 & 11.53 & 11.00 & 10.50 & 10.13 \\
\cmidrule{2-9}
 & \multirow{3}{*}{Qwen3-VL-8B-Instruct} & max & 16.05 & 27.59 & 25.56 & 28.56 & 25.70 & 25.95 \\
 &  & avg & 16.05 & 15.81 & 15.13 & 15.08 & 14.73 & 14.83 \\
 &  & min & 16.05 & 10.56 & 10.01 & 9.59 & 8.98 & 8.46 \\
\cmidrule{2-9}
 & \multirow{3}{*}{gemma-4-E4B-it} & max & 32.45 & 37.58 & 38.14 & 33.13 & 28.75 & 24.96 \\
 &  & avg & 32.45 & 23.56 & 21.62 & 19.13 & 17.54 & 16.74 \\
 &  & min & 32.45 & 13.67 & 12.72 & 12.11 & 11.18 & 10.75 \\
\bottomrule
    \end{tabular}
\end{table*}
\begin{table*}[hb]
    \setlength{\tabcolsep}{4pt}
    \begin{minipage}{1.1\textwidth}
    \begin{flushleft}
    Continuation of Table \ref{tab:res-id-complete}: \gls{cer} (\%) per configuration in the \gls{id} scenario for each model.
    \end{flushleft}
    \end{minipage}
    \centering
    \begin{tabular}{*{3}{l}*{6}{c}}
        \toprule
        \multirow{2}{*}{Collection} &\multirow{2}{*}{Model}&& \multicolumn{6}{c}{Context Size} \\
        \cmidrule{4-9}
         &  &  & 0 & 1 & 2 & 4 & 8 & 16 \\
    \midrule
\multirow{12}{*}{RIMES} & \multirow{3}{*}{Kimi-VL-A3B-Instruct} & max & 15.75 & 90.82 & 81.36 & 41.82 & 39.54 & 22.06 \\
 &  & avg & 15.75 & 15.41 & 12.69 & 10.23 & 9.00 & 7.57 \\
 &  & min & 15.75 & 4.07 & 3.75 & 3.44 & 3.26 & 3.16 \\
\cmidrule{2-9}
 & \multirow{3}{*}{Qwen2.5-VL-7B-Instruct} & max & 5.94 & 9.70 & 9.45 & 8.44 & 8.43 & 8.47 \\
 &  & avg & 5.94 & 5.23 & 4.91 & 4.61 & 4.52 & 4.43 \\
 &  & min & 5.94 & 2.98 & 2.78 & 2.62 & 2.57 & 2.51 \\
\cmidrule{2-9}
 & \multirow{3}{*}{Qwen3-VL-8B-Instruct} & max & 5.47 & 14.38 & 10.09 & 12.20 & 8.46 & 7.92 \\
 &  & avg & 5.47 & 5.37 & 4.97 & 4.93 & 4.61 & 4.34 \\
 &  & min & 5.47 & 3.04 & 2.60 & 2.77 & 2.54 & 2.46 \\
\cmidrule{2-9}
 & \multirow{3}{*}{gemma-4-E4B-it} & max & 8.09 & 11.86 & 11.72 & 11.28 & 11.27 & 10.96 \\
 &  & avg & 8.09 & 7.24 & 6.83 & 6.58 & 6.05 & 5.87 \\
 &  & min & 8.09 & 4.27 & 4.01 & 3.79 & 3.37 & 3.15 \\
\midrule
\multirow{12}{*}{SaintGall} & \multirow{3}{*}{Kimi-VL-A3B-Instruct} & max & 73.36 & 39.41 & 44.59 & 54.09 & 50.58 & 40.84 \\
 &  & avg & 73.36 & 21.13 & 20.76 & 21.67 & 22.35 & 22.15 \\
 &  & min & 73.36 & 13.87 & 13.37 & 13.36 & 14.74 & 15.54 \\
\cmidrule{2-9}
 & \multirow{3}{*}{Qwen2.5-VL-7B-Instruct} & max & 17.52 & 26.27 & 26.00 & 25.80 & 26.06 & 24.22 \\
 &  & avg & 17.52 & 16.82 & 17.09 & 17.10 & 17.21 & 17.29 \\
 &  & min & 17.52 & 11.79 & 11.18 & 10.78 & 11.18 & 11.56 \\
\cmidrule{2-9}
 & \multirow{3}{*}{Qwen3-VL-8B-Instruct} & max & 17.22 & 22.08 & 22.09 & 22.52 & 22.31 & 21.98 \\
 &  & avg & 17.22 & 16.91 & 16.69 & 17.24 & 17.44 & 17.14 \\
 &  & min & 17.22 & 13.24 & 12.50 & 12.32 & 12.82 & 12.69 \\
\cmidrule{2-9}
 & \multirow{3}{*}{gemma-4-E4B-it} & max & 23.80 & 24.91 & 22.77 & 22.45 & 27.63 & 23.91 \\
 &  & avg & 23.80 & 18.49 & 17.42 & 17.09 & 17.37 & 17.57 \\
 &  & min & 23.80 & 14.24 & 13.36 & 12.75 & 12.33 & 12.37 \\
\midrule
\multirow{12}{*}{Washington} & \multirow{3}{*}{Kimi-VL-A3B-Instruct} & max & 22.30 & 26.67 & 26.43 & 22.92 & 15.87 & 14.77 \\
 &  & avg & 22.30 & 8.87 & 8.69 & 8.30 & 7.63 & 7.36 \\
 &  & min & 22.30 & 4.89 & 4.63 & 4.39 & 4.30 & 4.06 \\
\cmidrule{2-9}
 & \multirow{3}{*}{Qwen2.5-VL-7B-Instruct} & max & 9.12 & 12.23 & 11.69 & 11.36 & 11.19 & 10.71 \\
 &  & avg & 9.12 & 8.11 & 7.86 & 7.55 & 7.22 & 6.91 \\
 &  & min & 9.12 & 5.30 & 4.88 & 4.53 & 4.10 & 3.73 \\
\cmidrule{2-9}
 & \multirow{3}{*}{Qwen3-VL-8B-Instruct} & max & 7.62 & 9.00 & 8.97 & 8.54 & 8.55 & 8.53 \\
 &  & avg & 7.62 & 6.11 & 5.92 & 5.74 & 5.61 & 5.45 \\
 &  & min & 7.62 & 4.10 & 3.87 & 3.61 & 3.54 & 3.25 \\
\cmidrule{2-9}
 & \multirow{3}{*}{gemma-4-E4B-it} & max & 20.90 & 26.24 & 24.81 & 19.62 & 24.04 & 18.55 \\
 &  & avg & 20.90 & 14.20 & 12.16 & 11.17 & 10.71 & 10.48 \\
 &  & min & 20.90 & 7.99 & 6.94 & 6.68 & 6.46 & 6.35 \\
\bottomrule
    \end{tabular}
\end{table*}
\begin{table*}[hb]
    \setlength{\tabcolsep}{4pt}
    \centering
    \caption{\gls{cer} (\%) per configuration in the \gls{cd} scenario for each model.}
    \label{tab:res-ood-complete}
    \begin{tabular}{*{3}{l}*{6}{c}}
        \toprule
        \multirow{2}{*}{Collection} &\multirow{2}{*}{Model}&& \multicolumn{6}{c}{Context Size} \\
        \cmidrule{4-9}
         &  &  & 0 & 1 & 2 & 4 & 8 & 16 \\
    \midrule
\multirow{12}{*}{IAM} & \multirow{3}{*}{Kimi-VL-A3B-Instruct} & max & 48.66 & 32.06 & 23.38 & 12.03 & 9.78 & 10.32 \\
 &  & avg & 48.66 & 9.23 & 7.29 & 5.93 & 5.56 & 5.47 \\
 &  & min & 48.66 & 4.56 & 4.45 & 4.46 & 4.43 & 4.37 \\
\cmidrule{2-9}
 & \multirow{3}{*}{Qwen2.5-VL-7B-Instruct} & max & 6.69 & 8.02 & 7.80 & 7.82 & 7.81 & 7.78 \\
 &  & avg & 6.69 & 6.37 & 6.28 & 6.29 & 6.27 & 6.22 \\
 &  & min & 6.69 & 5.37 & 5.33 & 5.36 & 5.37 & 5.33 \\
\cmidrule{2-9}
 & \multirow{3}{*}{Qwen3-VL-8B-Instruct} & max & 6.82 & 8.12 & 8.03 & 7.82 & 8.22 & 8.30 \\
 &  & avg & 6.82 & 6.58 & 6.60 & 6.58 & 6.61 & 6.63 \\
 &  & min & 6.82 & 5.48 & 5.60 & 5.72 & 5.72 & 5.72 \\
\cmidrule{2-9}
 & \multirow{3}{*}{gemma-4-E4B-it} & max & 8.56 & 11.54 & 11.58 & 11.50 & 12.16 & 11.99 \\
 &  & avg & 8.56 & 8.73 & 8.84 & 8.86 & 8.94 & 8.95 \\
 &  & min & 8.56 & 6.78 & 6.87 & 6.93 & 6.99 & 7.07 \\
\midrule
\multirow{12}{*}{ICFHR16} & \multirow{3}{*}{Kimi-VL-A3B-Instruct} & max & 896.86 & 1825.77 & 1782.70 & 1720.04 & 1816.87 & 1109.65 \\
 &  & avg & 896.86 & 190.18 & 189.65 & 191.60 & 194.58 & 144.57 \\
 &  & min & 896.86 & 62.10 & 62.29 & 63.41 & 63.77 & 64.02 \\
\cmidrule{2-9}
 & \multirow{3}{*}{Qwen2.5-VL-7B-Instruct} & max & 76.89 & 128.08 & 114.13 & 119.59 & 118.73 & 131.70 \\
 &  & avg & 76.89 & 75.74 & 75.04 & 75.33 & 74.85 & 76.08 \\
 &  & min & 76.89 & 59.79 & 60.43 & 60.95 & 61.03 & 61.94 \\
\cmidrule{2-9}
 & \multirow{3}{*}{Qwen3-VL-8B-Instruct} & max & 80.95 & 336.57 & 345.62 & 281.43 & 339.13 & 341.90 \\
 &  & avg & 80.95 & 94.34 & 95.00 & 93.61 & 98.28 & 100.66 \\
 &  & min & 80.95 & 63.72 & 63.99 & 65.23 & 65.79 & 65.95 \\
\cmidrule{2-9}
 & \multirow{3}{*}{gemma-4-E4B-it} & max & 457.82 & 1041.29 & 505.47 & 290.89 & 198.93 & 181.86 \\
 &  & avg & 457.82 & 286.75 & 152.38 & 115.76 & 96.87 & 93.61 \\
 &  & min & 457.82 & 87.87 & 75.48 & 71.28 & 68.84 & 68.86 \\
\midrule
\multirow{12}{*}{LAM} & \multirow{3}{*}{Kimi-VL-A3B-Instruct} & max & 50.26 & 204.67 & 208.58 & 202.75 & 187.44 & 179.38 \\
 &  & avg & 50.26 & 37.54 & 38.15 & 38.33 & 37.19 & 36.95 \\
 &  & min & 50.26 & 14.50 & 14.47 & 14.95 & 15.10 & 15.06 \\
\cmidrule{2-9}
 & \multirow{3}{*}{Qwen2.5-VL-7B-Instruct} & max & 23.34 & 82.88 & 69.48 & 65.29 & 56.23 & 59.64 \\
 &  & avg & 23.34 & 24.96 & 22.56 & 21.62 & 21.47 & 21.75 \\
 &  & min & 23.34 & 12.93 & 12.73 & 12.64 & 12.64 & 12.50 \\
\cmidrule{2-9}
 & \multirow{3}{*}{Qwen3-VL-8B-Instruct} & max & 16.05 & 35.46 & 32.15 & 28.41 & 27.81 & 29.30 \\
 &  & avg & 16.05 & 17.05 & 16.78 & 16.64 & 16.79 & 17.06 \\
 &  & min & 16.05 & 11.39 & 11.22 & 11.58 & 11.88 & 12.10 \\
\cmidrule{2-9}
 & \multirow{3}{*}{gemma-4-E4B-it} & max & 32.45 & 54.54 & 52.27 & 43.32 & 41.00 & 35.51 \\
 &  & avg & 32.45 & 26.36 & 25.36 & 23.21 & 21.36 & 20.97 \\
 &  & min & 32.45 & 14.19 & 13.91 & 14.01 & 13.53 & 13.67 \\
    \bottomrule
    \end{tabular}
\end{table*}
\begin{table*}[hb]
    \setlength{\tabcolsep}{4pt}
    \begin{minipage}{1.1\textwidth}
    \begin{flushleft}
        Continuation of Table \ref{tab:res-ood-complete}: \gls{cer} (\%) per configuration in the \gls{cd} scenario for each model.
    \end{flushleft}
    \end{minipage}
    \centering
    \begin{tabular}{*{3}{l}*{6}{c}}
        \toprule
        \multirow{2}{*}{Collection} &\multirow{2}{*}{Model}&& \multicolumn{6}{c}{Context Size} \\
        \cmidrule{4-9}
         &  &  & 0 & 1 & 2 & 4 & 8 & 16 \\
    \midrule
\multirow{12}{*}{RIMES} & \multirow{3}{*}{Kimi-VL-A3B-Instruct} & max & 15.75 & 103.59 & 70.35 & 57.13 & 33.46 & 28.97 \\
 &  & avg & 15.75 & 17.59 & 14.95 & 12.84 & 11.36 & 10.45 \\
 &  & min & 15.75 & 4.42 & 4.37 & 4.72 & 4.79 & 4.80 \\
\cmidrule{2-9}
 & \multirow{3}{*}{Qwen2.5-VL-7B-Instruct} & max & 5.94 & 9.95 & 8.81 & 8.58 & 9.19 & 9.26 \\
 &  & avg & 5.94 & 5.38 & 5.19 & 5.07 & 5.19 & 5.33 \\
 &  & min & 5.94 & 3.29 & 3.29 & 3.12 & 3.28 & 3.18 \\
\cmidrule{2-9}
 & \multirow{3}{*}{Qwen3-VL-8B-Instruct} & max & 5.47 & 8.36 & 8.23 & 7.41 & 7.53 & 7.68 \\
 &  & avg & 5.47 & 5.19 & 5.07 & 5.08 & 5.15 & 5.22 \\
 &  & min & 5.47 & 3.38 & 3.41 & 3.57 & 3.41 & 3.44 \\
\cmidrule{2-9}
 & \multirow{3}{*}{gemma-4-E4B-it} & max & 8.09 & 13.96 & 12.94 & 12.74 & 12.53 & 12.39 \\
 &  & avg & 8.09 & 7.50 & 7.67 & 7.69 & 7.54 & 7.49 \\
 &  & min & 8.09 & 4.69 & 4.48 & 4.64 & 4.45 & 4.63 \\
\midrule
\multirow{12}{*}{SaintGall} & \multirow{3}{*}{Kimi-VL-A3B-Instruct} & max & 73.36 & 152.93 & 117.99 & 87.47 & 98.86 & 73.62 \\
 &  & avg & 73.36 & 29.20 & 28.11 & 27.48 & 28.17 & 26.58 \\
 &  & min & 73.36 & 15.30 & 15.60 & 16.36 & 16.83 & 17.05 \\
\cmidrule{2-9}
 & \multirow{3}{*}{Qwen2.5-VL-7B-Instruct} & max & 17.52 & 27.58 & 26.96 & 27.80 & 28.81 & 29.31 \\
 &  & avg & 17.52 & 17.99 & 18.00 & 17.99 & 18.05 & 17.97 \\
 &  & min & 17.52 & 13.15 & 12.94 & 12.93 & 12.94 & 12.97 \\
\cmidrule{2-9}
 & \multirow{3}{*}{Qwen3-VL-8B-Instruct} & max & 17.22 & 21.84 & 21.98 & 22.03 & 22.14 & 21.94 \\
 &  & avg & 17.22 & 17.49 & 17.46 & 17.45 & 17.43 & 17.33 \\
 &  & min & 17.22 & 14.03 & 13.85 & 13.92 & 13.94 & 13.96 \\
\cmidrule{2-9}
 & \multirow{3}{*}{gemma-4-E4B-it} & max & 23.80 & 28.37 & 26.88 & 26.09 & 26.51 & 26.57 \\
 &  & avg & 23.80 & 20.36 & 19.62 & 19.36 & 19.43 & 19.63 \\
 &  & min & 23.80 & 15.54 & 15.11 & 15.13 & 15.18 & 15.36 \\
\midrule
\multirow{12}{*}{Washington} & \multirow{3}{*}{Kimi-VL-A3B-Instruct} & max & 22.30 & 97.05 & 45.49 & 23.22 & 32.17 & 22.15 \\
 &  & avg & 22.30 & 15.56 & 10.52 & 9.68 & 9.96 & 9.64 \\
 &  & min & 22.30 & 5.23 & 5.20 & 5.55 & 5.58 & 5.89 \\
\cmidrule{2-9}
 & \multirow{3}{*}{Qwen2.5-VL-7B-Instruct} & max & 9.12 & 12.89 & 12.57 & 12.06 & 12.06 & 12.20 \\
 &  & avg & 9.12 & 8.70 & 8.65 & 8.57 & 8.43 & 8.45 \\
 &  & min & 9.12 & 6.05 & 6.09 & 6.05 & 5.96 & 6.01 \\
\cmidrule{2-9}
 & \multirow{3}{*}{Qwen3-VL-8B-Instruct} & max & 7.62 & 9.08 & 8.85 & 8.52 & 8.43 & 8.58 \\
 &  & avg & 7.62 & 6.43 & 6.39 & 6.31 & 6.31 & 6.35 \\
 &  & min & 7.62 & 4.63 & 4.67 & 4.74 & 4.79 & 4.78 \\
\cmidrule{2-9}
 & \multirow{3}{*}{gemma-4-E4B-it} & max & 20.90 & 32.56 & 30.62 & 27.18 & 29.50 & 27.37 \\
 &  & avg & 20.90 & 16.28 & 13.93 & 14.59 & 14.53 & 14.52 \\
 &  & min & 20.90 & 8.49 & 8.37 & 8.21 & 8.15 & 8.25 \\
    \bottomrule
    \end{tabular}
\end{table*}
}

\end{document}